# AEGIS: A real-time multimodal augmented reality computer vision based system to assist facial expression recognition for individuals with autism spectrum disorder

J. R. H. Lee[1,2], A. Wong [1,2]

[1]University of Waterloo, [2]Vision and Image Processing Lab

## INTRODUCTION

When interacting with others in society, understanding social cues and the flow of conversation is something that happens naturally for most people. However, individuals living with Autism Spectrum Disorder (ASD) can have an increased difficulty in interpreting and understanding facial expressions and their corresponding emotions [1,2]. This deficiency sometimes causes troubles integrating into society, as well as problems creating and sustaining meaningful, positive relationships [1], leading to a higher prevalence of depression and loneliness. However, studies have shown that after the use of assistive technology (AT) software, participants with ASD showed improvement on facial emotion recognition for emotions shown in the software, as well as emotions not included in the software [2]. This finding leads us to believe that the existence of a real-time expression classification system could help alleviate some of these issues faced by peoples with ASD, and thus we propose AEGIS (Augmented-reality Expression Guided Interpretation System), designed in order to assist these individuals learn to better identify expressions and improve their social experiences. AEGIS is a multimodal augmented reality (AR) assistive technology system deployable on a wide range of user devices including tablets, smartphones, video conference systems, and smartglasses, showcasing its extreme flexibility in a variety of use cases, to allow integration into daily life with ease. AEGIS leverages the use of computer vision and deep convolutional neural networks in order to achieve accurate yet real-time performance, granting the user a seamless, intuitive experience that can assist them in their societal interactions.

Facial expression classification (FEC) is a highly studied field, due to its wide range of applications. We plan on using our system as an assistive technology, but automatic expression classification can also be used for driver state monitoring, analyzing customer reactions in marketing, security, ambient interfaces, and empathetic tutoring [3,4]. In literature, most studies tend to focus on either offline video classification, that is, taking an entire video sequence as input and providing a *single* label as output, or single frame classification, which is classifying *each frame* in the video [3,5,6]. We take an intermediary approach, combining the fast inference speeds from a smaller number of frames with the accuracy of sequential information, and use these video sub-sequences instead as our inputs. The problem of facial expression classification is a difficult one, and is something that even humans have trouble with, due to the wide range of emotions we are capable of expressing, as well as the fact that human facial expressions are of a dynamic nature [5], consisting of onset, peak, and offset phases. Thus, to simplify the problem, most studies end up reducing the problem subspace into the six basic universal emotions (anger, disgust, fear, happiness, sadness, and surprise) [5] as well as neutral. To deal with the dynamic nature of expressions, temporal information can be used to capture the smooth transition between expressions as well as their transient nuances, and thus these temporal-based methods can result in higher accuracy when compared to single frame classification, which can generally achieve high classification accuracy but only when the input is a peak phase expression. Many methods exist in literature that can handle temporal-based inputs [4,5], but the majority of them do not run in real-time, which is a key requirement when assisting users with ASD, as we intend to provide them with a seamless experience that does not interfere with their daily lives. Typically, FEC approaches involve the use of some machine learning algorithm, such as Support Vector Machines [3,7], in combination with some set of features, either handcrafted (e.g. Local Binary Patterns [3], Facial Action Coding System (FACS) [8]), or deep (e.g. CNN [5]), but due to the additional preprocessing that must be performed to use features such as FACS, as well as the complexity of the task, we chose to use a deep neural network due to its high performance and low latency capabilities when operating on nonlinear problems.

## APPROACH

AEGIS uses a device with a screen and a video streaming camera, which could range from a handheld device such as a mobile phone or tablets, to wearable technology such as smartglasses. Our software would be installed on the device, and would run in real-time. The user would face the camera towards the desired area, which would be the faces of the people they are planning on talking to, while watching the screen of the device themselves. As shown in Figure 1, AEGIS takes in a streaming video camera source as input and processes each real-world frame before automatically passing them to our novel deep convolutional time windowed neural network (TimeConvNet [9]). Within the model, sets of learned convolutional filters leverage both spatial and temporal information in order to provide an accurate expression prediction. The original real-world camera frame is then augmented with the corresponding emoji, and finally shown to the user on the device screen. This entire process takes milliseconds to complete, allowing for a seamless and immersive experience where the user sees augmented real-world frames with emojis overlaid on top of each person in real-time.

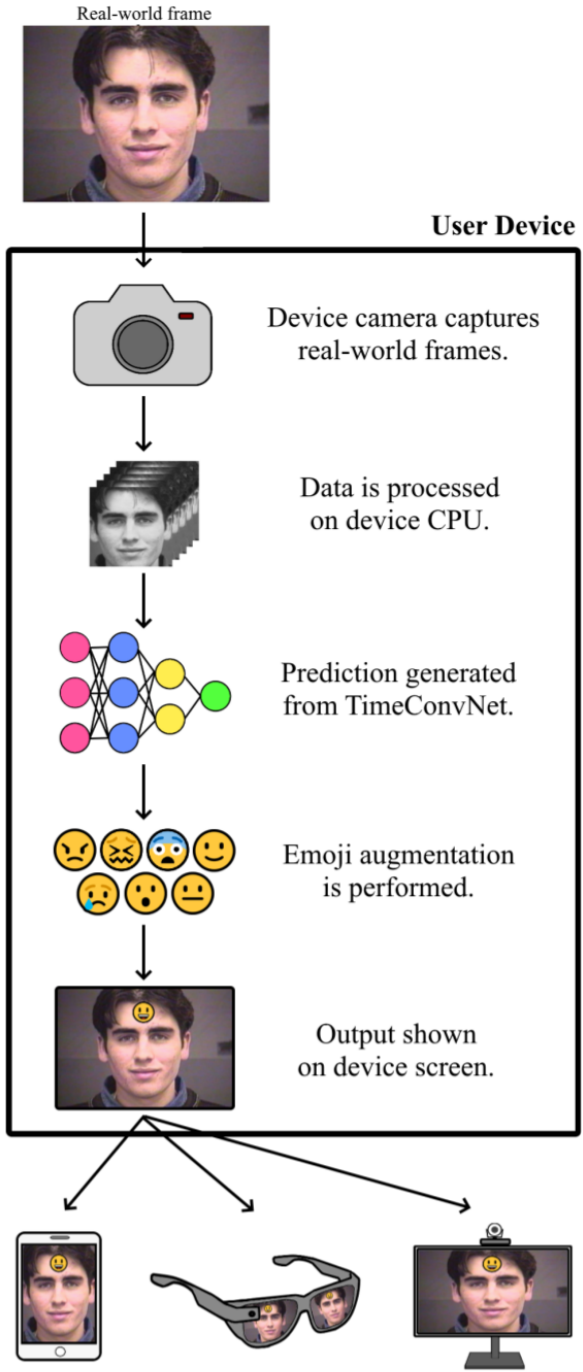

Figure 1. System overview of AEGIS. Given a streaming video camera as input, each real-world frame is processed on the device CPU, then passed to our novel TimeConvNet. Emoji augmentation is performed based on the network prediction, and the output is shown on the user device. AEGIS is deployable on a variety of devices, including smartphones, smartglasses, and video conference systems.

**Visualization Choices**

Naturally, the way we present the information to the user is a key factor in how well they understand the given message. In order to provide the user with an immersive experience, the expression information must be presented in a way such that the user can understand the information unambiguously and without thought. The work presented in [1] reports that 50 percent of individuals diagnosed with ASD are highly visual-oriented and possess strong visual-spatial abilities, and additional studies have also found that these individuals respond much better to visual based stimuli when compared to auditory based stimuli. Inspired by these findings, and motivated by the speed and clarity requirements, we designed our visualization choices to best suit these visual-based needs.

The first option we thought of was using a coloured outline, which would essentially show each person in the frame with a glowing outline of a certain colour, based on their predicted emotion. However, we realized that it might be difficult to associate certain colours to certain emotions, as it may vary for each individual. Also, there might not exist a colour that most accurately describes an emotion - for example, anger might be thought of as red, but what about emotions such as surprise or happiness? Thus, we decided on using emojis as our visualization choice, where we would show an emoji floating over each person's head. The reasoning behind this choice was because of their growing popularity among people, both young and old, and also due to their expressiveness and ability to immediately convey certain tones. This came with its own set of challenges, but we managed to find emojis that corresponded well to each of the seven emotions we planned on showing.

For those individuals who prefer a text-based system, a text overlay with a prediction confidence percentage is also a possible design. In literature, some real-time FEC approaches tend to show their prediction results with labelled bounding boxes around the individual faces [6,10], but for visual-oriented users, this method is suboptimal, as an emoji is able to convey a message tone much faster than a word can, and are generally much more intuitive. Thus, for initial prototypes of the system, we chose to use the emoji-based visual-oriented approach.

**Neural Network Design**

As we are aiming for a real-time system, we must choose a good balance between performance accuracy and inference speed. We use a novel deep time windowed convolutional neural network design which we call TimeConvNets [9]. Given a streaming video sequence, if there is a face detected in the frame, we process said frame by cropping it to the facial bounding box and then resizing to a size of 48x48 pixels. We then add these processed frames into a first-in-first-out queue of length $t$. By doing so, we create a dynamic $t$-channel video sub-sequence, which we then stack together in the channel dimension (creating an input of shape 48x48x$t$), and provide this stack as the input to our TimeConvNet system, where a set of learned convolutional filters within the convolutional spatiotemporal encoding layer captures both the spatial and temporal attributes of the time window. This time windowed method allows us to capture the transient nuances of dynamic facial expressions without the use of computationally complex 3D convolutions or needing an entire video sequence. The spatiotemporal encoding layer is fed into a backbone convolutional neural network architecture, where progressively higher levels of abstraction is performed for improved discrimination amongst the facial expression categories. Finally, a softmax layer determines the prediction, which is one of seven classes (the six basic universal emotions, plus neutral).

We tested the TimeConvNet architecture with a variety of backbone networks, such as mini-Xception [6], ResNet20 [11], and MobileNetV2 [12], all chosen due to their reputations as widely used compact network designs that offer a strong balance between accuracy and efficiency. Due to the low latency requirement for AEGIS, as users would want expression feedback in real-time for a seamless experience, we ultimately decided on the ResNet20 backbone architecture as it provided the best balance compared to the other models. We found that by using the ResNet20 TimeConvNet in our system, we could achieve a frame rate of roughly 25 FPS [13], which can allow for smooth social interactions.

**Hardware Choices**

The requirements of the device used in AEGIS was one that was simple to use, easy to set up, and also non-obstructive towards the everyday life of the user. Camera input was needed, and a screen was required to display the output. These simple requirements allow the AEGIS to be deployable on a variety of devices, including tablets and the increasingly popular smartphone. The system is also usable on video conference systems and smartglasses, to account for the fact that certain interactions may be negatively impacted due to people feeling uncomfortable having a camera pointed at them during conversation, and also to allow for a wider variety of use cases including in business social interactions. For now, initial prototypes will be made to run on smartphones, but deployment to these other platforms is possible in the future due to the extreme flexibility of the system.

**IMPLICATIONS**

Once deployed, AEGIS will be easily accessible and usable by anyone owning a smartphone device. The system comes at virtually no cost, and as long as the user owns the appropriate hardware device, they will be able to access and use the application. With the use of our multimodal system, these individuals with ASD will no longer be required to guess the emotional states of other people, and can easily interpret their expressions via the non-intrusive yet simple to understand emojis. Further research will need to be done to determine if users with ASD actually benefit from using our system, but studies such as [2] which employ similar assistive technologies have shown that improved emotion recognition skills can be achieved after the use of their system. In [2] specifically, when compared against pretest performances, participant scores were on average ~19% higher on the task where emotions were inferred from facial expression video clips, and on average ~9% higher on the task where emotions were inferred from voice audio clips. These promising results lead us to believe that AEGIS can achieve comparable improvements due to the similarities between the assistive technologies.

However, one problem is that users may be unwilling to use it in a social setting at first, as they will need to have their phone cameras on and pointed towards anyone they are speaking to. Also, social interactions may be negatively influenced due to the other party being unable or unwilling to accommodate the system. Thus, we propose that the system be initially used for training purposes, in an environment where all parties are comfortable and familiar with the system and its use, such as at home. Through use of the system, those living with ASD may be able to learn and interpret social cues firstly for people close to them, and then be able to generalize to a wider range of interactions in society. Once AEGIS is deployed on video conference systems and smartglasses, individuals will also be able to use our system in more unfamiliar settings, allowing them to gain confidence in societal interaction and thus improve their social lives.

**CONCLUSION**

This paper presented the development of AEGIS, a novel multimodal augmented reality assistive technology system designed to help individuals with ASD with the detection and interpretation of facial expressions in social settings. The first iteration of the system is planned to run on a smartphone device, and overlay expression information via emojis on top of each real-world camera frame, in real-time. We propose that the system be initially used in a home setting for training purposes, and then allow the user to generalize to experiences in society. Future work involves deploying the system to other devices such as tablets and smartglasses, and validating the benefits of using our system.

**ACKNOWLEDGEMENT**

The authors would like to thank NSERC, Canada Research Chairs program, and Microsoft for their support.